\documentclass[letterpaper, 10pt, conference]{ieeeconf}  %

\usepackage{booktabs}
\usepackage{multirow}
\usepackage{subcaption}
\usepackage{xspace}

\usepackage{graphics} %
\usepackage{times} %
\usepackage{amsmath} %
\usepackage{amssymb}  %
\usepackage{todonotes}
\usepackage{booktabs}
\usepackage{hyperref}
\usepackage{cleveref}
\usepackage{subcaption}
\usepackage{bm}
\usepackage{algorithm}
\usepackage{soul}

\usepackage{bigfoot}
\DeclareNewFootnote{ANote}[fnsymbol]
\makeatletter
\def\thanksANote#1{%
  \protected@xdef\@thanks{%
        \protect\footnotetextANote[\the \c@footnoteANote]{$^*$#1}}%
}
\makeatother

\usepackage{amsmath,amsfonts,bm}

\def\eqref#1{equation~\ref{#1}}

\def\1{\bm{1}}

\DeclareMathAlphabet{\mathsfit}{\encodingdefault}{\sfdefault}{m}{sl}
\SetMathAlphabet{\mathsfit}{bold}{\encodingdefault}{\sfdefault}{bx}{n}

\def\gA{{\mathcal{A}}}

\def\gD{{\mathcal{D}}}

\def\gG{{\mathcal{G}}}

\def\gI{{\mathcal{I}}}

\def\gO{{\mathcal{O}}}
\def\gP{{\mathcal{P}}}

\def\gi{{\mathcal{I}\textit{mg}}}

\usepackage{amsmath} %
\usepackage{amssymb}  %

\usepackage{booktabs}
\usepackage{hyperref}
\usepackage{dblfloatfix}
\usepackage{graphicx} 
\usepackage{caption}
\usepackage{subcaption}

\newcommand{\proc}[1]{\textsc{#1}}

\usepackage{xcolor}

\definecolor{LightGreen}{RGB}{226, 246, 211}
\definecolor{DarkGreen}{RGB}{184, 233, 148}

\definecolor{MyDarkBlue}{rgb}{0,0.08,0.8}
\definecolor{MyDarkGreen}{RGB}{45,155,45}
\definecolor{MyDarkRed}{rgb}{0.8,0.02,0.02}
\definecolor{MyDarkOrange}{rgb}{0.40,0.2,0.02}
\definecolor{MyPurple}{RGB}{111,0,255}
\definecolor{MyRed}{rgb}{0.8,0.0,0.0}
\definecolor{MyGold}{rgb}{0.75,0.6,0.12}
\definecolor{MyDarkgray}{rgb}{0.66, 0.66, 0.66}


\usepackage{xcolor}
\usepackage{algorithm}
\usepackage[noend]{algpseudocode}
\usepackage[export]{adjustbox}
\algnewcommand\algorithmicdeclare{\textbf{Assume:}}
\algnewcommand\Declare{\item[\algorithmicdeclare]}

\usepackage{listings}
\let\labelindent\relax
\usepackage{enumitem}

\usepackage{pifont} %

\usepackage{algorithm}
\usepackage[noend]{algpseudocode}

\usepackage{framed}
\definecolor{shadecolor}{gray}{0.95}

\usepackage{xparse}
\usepackage{etoolbox}
\usepackage{setspace}

\usepackage{tcolorbox}
\usepackage{lipsum}
\tcbuselibrary{skins,breakable}
\usetikzlibrary{shadings,shadows}
    {\endtcolorbox}

\definecolor{ColorAskVLM}{RGB}{203, 242, 220}
\definecolor{ColorVLMAnswer}{RGB}{204, 229, 245}
\definecolor{ColorComment}{RGB}{116, 125, 140}
\definecolor{ColorKeyword}{RGB}{87, 96, 111}
\usepackage{listings}
\usepackage{changepage}
\lstdefinestyle{askvlm}{
  rulecolor=\color{ColorAskVLM},
  backgroundcolor=\color{ColorAskVLM},
}
\lstdefinestyle{vlmanswer}{
  rulecolor=\color{ColorVLMAnswer},
  backgroundcolor=\color{ColorVLMAnswer},
}

\usepackage{setspace}

\usepackage[font=small]{caption}

\newcommand{\ignore}[1]{}

\makeatletter
\DeclareRobustCommand\onedot{\futurelet\@let@token\@onedot}
\def\@onedot{\ifx\@let@token.\else.\null\fi\xspace}

\makeatother

\definecolor{MyDarkBlue}{rgb}{0,0.08,1}
\definecolor{MyDarkGreen}{rgb}{0.02,0.6,0.02}
\definecolor{MyDarkRed}{rgb}{0.8,0.02,0.02}
\definecolor{MyDarkOrange}{rgb}{0.40,0.2,0.02}
\definecolor{MyPurple}{RGB}{111,0,255}
\definecolor{MyRed}{rgb}{1.0,0.0,0.0}
\definecolor{MyGold}{rgb}{0.75,0.6,0.12}
\definecolor{MyDarkgray}{rgb}{0.66, 0.66, 0.66}
\definecolor{YangRed}{RGB}{231, 76, 60}
\definecolor{YangGreen}{RGB}{39, 174, 96} 
\definecolor{YangBlue}{RGB}{52, 152, 219}
\definecolor{YangOrange}{RGB}{230, 126, 3}
\definecolor{YangPurple}{RGB}{142, 68, 173}

\newcommand{\ours}{VLM-TAMP\xspace}

\newif\ifpropositionfirstitem
\propositionfirstitemtrue

\title{\LARGE \bf
Guiding Long-Horizon Task and Motion Planning \\ with Vision Language Models
}

\author{%
Zhutian Yang$^{1, 2*}$%
\thanksANote{Work done during a part-time internship at NVIDIA Research. Corresponding author: Zhutian Yang (\url{ztyang@mit.edu}).} 
\quad Caelan Garrett$^{2}$
\quad Dieter Fox$^{2}$\quad
Tomás Lozano-Pérez$^{1}$\quad
Leslie Pack Kaelbling$^{1}$\\
$^{1}$Massachusetts Institute of Technology, $^{2}$NVIDIA Research
}

\begin{document}

\setcounter{figure}{1}
\makeatletter
\let\@oldmaketitle\@maketitle%
\renewcommand{\@maketitle}{\@oldmaketitle%
\vspace{4pt}
  \begin{center}
    \includegraphics[width=0.99\linewidth]{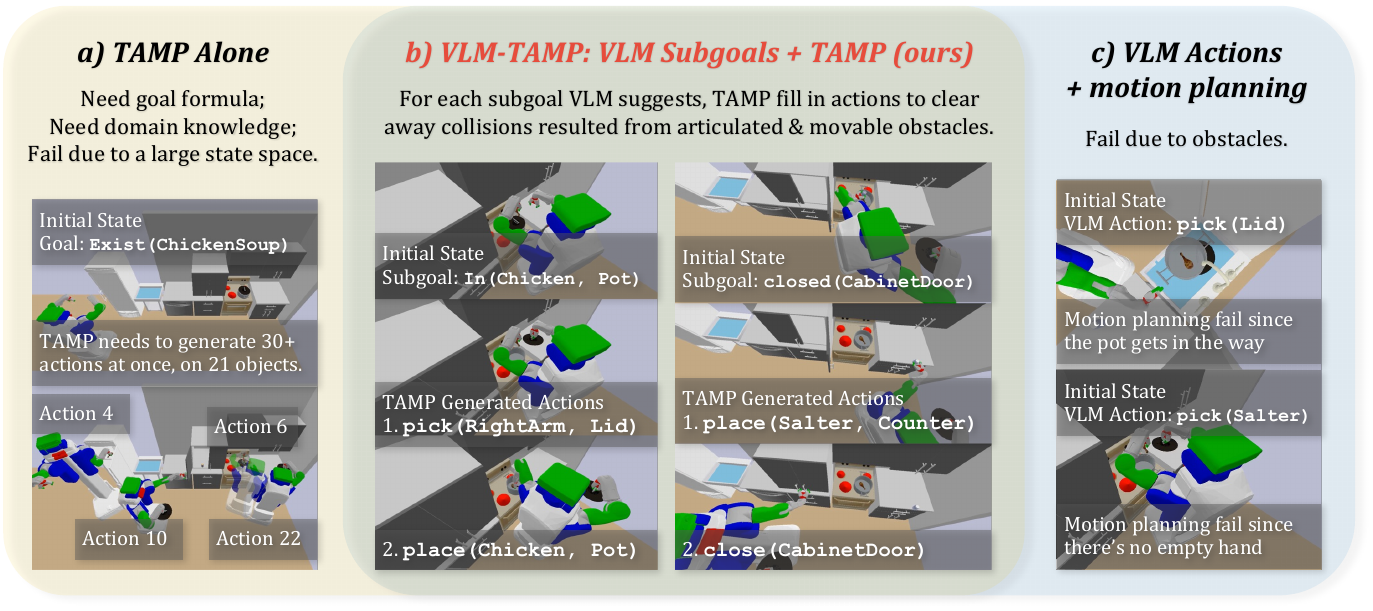} 
    \label{fig:spectrum}
  \end{center}
  \vspace{-8pt}
  {\small
  \begin{singlespace}
  Fig.~\thefigure:~ %
  Our approach \ours overcomes the pitfalls of using TAMP alone and VLM task {\em then} motion planning when solving long-horizon robot manipulation problems.
  \textbf{a)} Pure TAMP fails 
  when there are large state spaces and a long-horizon goals. \textbf{c)} VLMs fail at %
  geometric reasoning by predicting %
  actions that cannot be safely refined with motion planning. \textbf{b)} \ours uses VLM to take in a natural language goal, generate subgoals, solve a sequence of smaller problems that respect all geometric constraints using TAMP.
  \end{singlespace}
  }
  \vspace{-8pt}
  \medskip}%
\makeatother
\maketitle

\begin{abstract}
Vision-Language Models (VLM) can generate plausible high-level plans when prompted with a goal, the context, an image of the scene, and any planning constraints. However, there is no guarantee that the predicted actions are geometrically and kinematically feasible for a particular robot embodiment. As a result, many prerequisite steps such as opening drawers to access objects are often omitted in their plans. Robot task and motion planners can generate motion trajectories that respect the geometric feasibility of actions and insert physically necessary actions, but do not scale to everyday problems that require common-sense knowledge and involve large state spaces comprised of many variables. 
We propose \ours, a hierarchical planning algorithm that leverages a VLM to generate goth semantically-meaningful and horizon-reducing intermediate subgoals that guide a task and motion planner.
When a subgoal or action cannot be refined, the VLM is queried again for replanning. 
We evaluate \ours{} on kitchen tasks where a robot must accomplish cooking goals that require performing 30-50 actions in sequence and interacting with up to 21 objects. 
\ours substantially outperforms baselines that rigidly and independently execute VLM-generated action sequences, both in terms of success rates (50 to 100\% versus 0\%) and average task completion percentage (72 to 100\% versus 15 to 45\%). See project site \url{https://zt-yang.github.io/vlm-tamp-robot/} for more information.
\end{abstract}
\section{Introduction}

Large Language Models (LLMs) contain an enormous amount of common-sense and cultural knowledge through training on internet-scale datasets~\cite{bommasani2021opportunities,achiam2023gpt}.  They can suggest high-level courses of action for solving almost any problem, ranging from selling your house to making a meal.  
Vision Language Models (VLMs) extend the capabilities of LLMs, by conditioning on an input image, allowing problems to be described both textually and via one or more pictures.

\begin{figure}[tb]
    \centering
    \begin{subfigure}[b]{0.17\textwidth}
        \includegraphics[width=\textwidth]{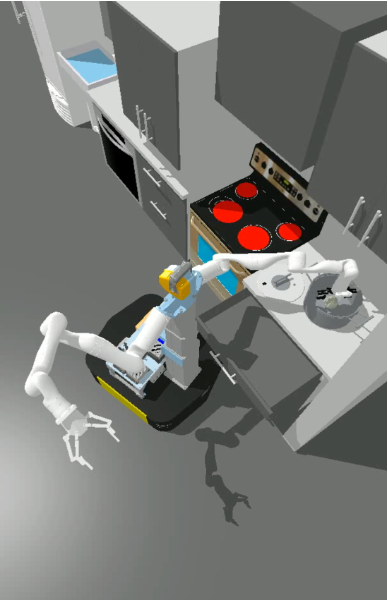}
        \caption{Dual-arm Rummy directly picks and places the cabbage as its arms are long enough to reach far.}
    \end{subfigure} \hfill 
    \begin{subfigure}[b]{0.143\textwidth}
        \includegraphics[width=\textwidth]{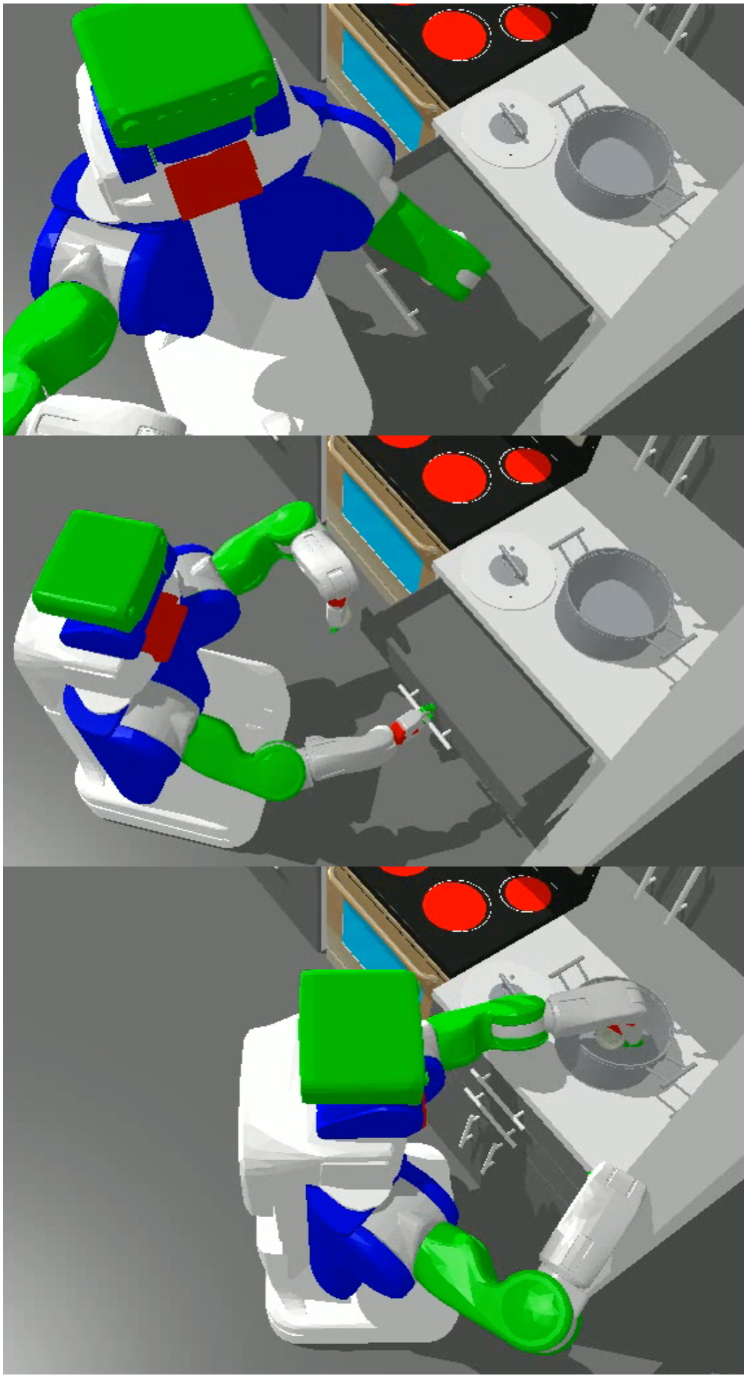}
        \caption{Dual-arm PR2 first closes the drawer to make space for reaching the pot.}
    \end{subfigure} \hfill 
    \begin{subfigure}[b]{0.1475\textwidth}
        \includegraphics[width=\textwidth]{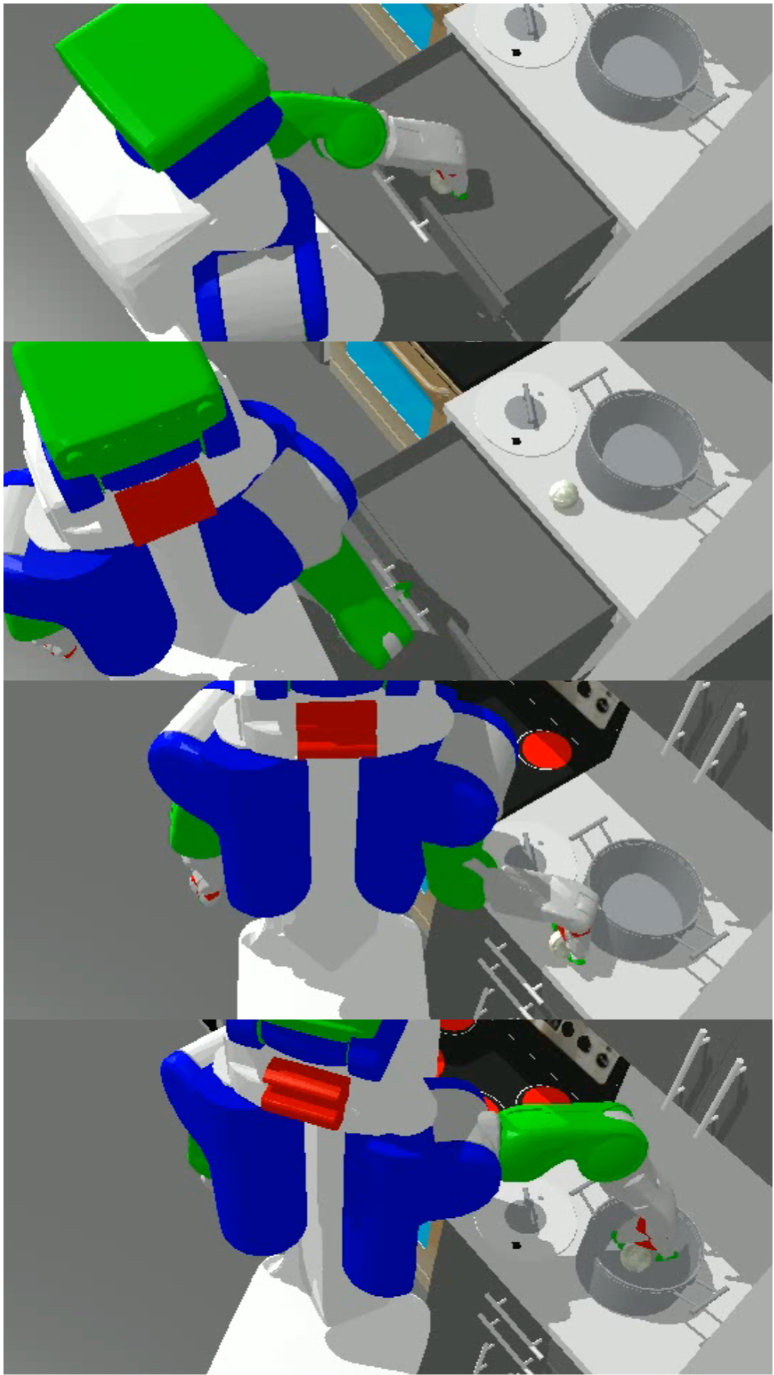}
        \caption{Single-arm PR2 first puts the cabbages aside before closing the drawer.}
    \end{subfigure}
    \caption{\small Example trajectories of different robots achieving the same goal of having the cabbage in the pot, where the cabbage is placed in a drawer and the pot is hard to reach. While the VLM may not be able to generate feasible action plans based on text and image description of the scene, TAMP can find the shortest feasible task plans that move obstacles if necessary and respect the kinematic constraints of the robot and other articulated objects.}
    \label{fig:embodiment}
    \vspace{-8pt}
\end{figure}

However, neither method is capable of detailed geometric reasoning:  they don't understand whether a particular robot with particular kinematics can reach something or whether it's possible to fit two particular pans into the oven simultaneously.  Furthermore, because they are trained on human-generated text, which usually only contains the most important aspects of a plan but leaves unstated many steps that are obvious to a human (you have to open the fridge to get the eggs, it's a good idea to close it again, you should extract the egg from the shell before adding it to your cake, etc.) The plans they generate are error-prone and partial. Besides, the same plan may be feasible for some robot embodiment but not the others depending on the robot reachability, as shown in an example in Figure~\ref{fig:embodiment}.

In the robotics research community, task and motion planning (TAMP)~\cite{Garrett2021} methods can solve complex long-horizon manipulation problems.  These planners are sound and (semi) complete:  that is, given an accurate model of the domain, and a problem to be solved that satisfies some common assumptions, they are {\em guaranteed} to eventually produce a detailed plan, at the level of robot joint commands, that achieve the goal, if such a plan exists. Two weakness of TAMP approaches are how they 1) handle open-world {\em semantics}, they can only address the geometric and kinematic aspects of the problem (they can plan to put a chicken in a pot, but don't know what it means to make tasty soup), and 2) scale {\em computationally}, the solution time explodes with complexity of the problem (the number of objects that need to be manipulated and the length of the solution plan).

In this paper, we present \ours{}, a system that combines the great strengths of VLMs and TAMP (Figure~1). It uses a VLM to suggest a sequence of {\em subgoals} such that achieving them in order could achieve a higher-level common-sense goal (like making tasty soup).  These subgoals are then solved by a TAMP algorithm, which can fill in missing geometric details, insert new steps (e.g., opening a cupboard) if needed, and approximately detect infeasibility.  
The system executes the actions after solving each subgoal. 
In case one subgoal suggested by the VLM cannot be grounded in the planning domain or is geometrically infeasible, \ours{} reprompts the VLM to generate a new subgoal sequence given the current state of the world and objects that are collided in simulation during the last failed planning process.

We evaluate \ours on two robot embodiments, solving cooking problems in procedurally generated kitchen scenes. The problems consist of making chicken soup with different scene initial conditions, requiring task plans that range from 30 to 50 actions.  We compare \ours{} extensively to a more common strategy, in which the VLM is asked to produce a sequence of {\em actions}, which are then directly refined by sampling the necessary continuous parameters, and executed in the world.  We find that the ability of \ours{} to make up for deficiencies in the high-level plan makes it much superior.  Our experiments show that, in a challenging set of very long-horizon problems, \ours succeeds 50 to 100\% of the time, where as the baseline, a naive VLM predicting actions + motion planning approach never succeeds. When we count task progress, the baseline completes on average only 15 to 45\% of the subproblems before failing to refine an action the VLM suggested, even given chances to reprompt VLM with collision information. In comparison, ours, which asks VLM for subgoals, visibly benefit from reprompting, with task success increasing by 47 to 55\% on the hardest problems with 21 planning objects in a layout that resembles real-world kitchens.

\section{Related Work}
\label{sec:related}

LLMs and VLMs have been used to translate natural language task description to formal languages \cite{liu2023llm+,ding2023task,chen2024autotamp}, which a model-based planner can consume and solve. 
They have been prompted to generate high-level action sequences that are then implemented by pre-trained skill policies \cite{ahn2022can, huang2023grounded}. They have also been used to generate code for calling robot motion primitives \cite{singh2023progprompt,liang2023code} and guiding trajectory generation \cite{huang2023voxposer}. These methods do not explicitly handle the problem of geometrically infeasible action sequences.
 
To correct the infeasible task plans generated by LLMs due to obstacles or partial observability, \cite{zhang2023grounding}, \cite{skreta2024replan}, \cite{wang2024llm}, and \cite{joublin2023copal} provide replanning abilities by prompting the VLM to adapt robot actions when the initial plan fails to achieve the desired goal. These methods showed improved success rate on small-scale table-top rearrangement tasks. But using the VLM or LLM for geometric reasoning assumes that the infeasibility can be described with a given text template ``some object is blocking the goal object'', which does not really scale to larger environments, especially involving mobile manipulators.
For example, when the goal is to pick up a pepper shaker from a small cabinet and the robot has only one arm, the actions to resolve the geometric infeasibility would involve putting down the object in the hand and opening both cabinet doors. While a VLM is unlikely to come up with all these actions, a TAMP planner can.

\cite{guo2023doremi} prevents LLMs from proposing actions that violate geometric constraints by querying the LLMs to generate the constraints, such as that the robot is holding an object and thus unable to pick up another object. Designing such queries is equivalent to providing the planning domain knowledge for checking violations. 
To incorporate geometric feasibility reasoning directly into LLM and VLM planning, \cite{lin2023text2motion} proposed a shooting-based strategy, where the LLM proposes K task plans, then geometric feasibility planning is carried out to find continuous parameters, the sequences’ success probability and predicted future states. Their method considers the geometric dependencies spanning the whole action sequence, but the planner depends on Q-functions that are specific to the robot embodiment and scene layout. Furthermore, the method doesn't enable local geometric (in)feasibility to guide which task plans to try next.

Our work differs from all these methods in that we use an LLM to generate intermediate goals instead of actions~\cite{xie2023translating}, and deploy TAMP planner to achieve them.
This enables \ours{} to substantially modify an LLM's suggested partial plan by adding steps and by solving continuous parameters that ensure geometric and kinematic feasibility.

\section{Method}
\label{sec:methods}

We present \ours, a planning algorithm that uses a VLM to break down a long-horizon manipulation planning problem (defined in Section~\ref{sec:formulation}) into a sequence of smaller ones (Section~\ref{sec:vlm}), which a TAMP planner solves in sequence to satisfy geometric feasibility (Section~\ref{sec:tamp}). \ours also contains a replanning mechanism that deals with mistakes in VLM goal translation, infeasible task plans, and TAMP planning failures (Section~\ref{sec:replanning}).

\subsection{Problem Formulation}\label{sec:formulation}

\begin{figure}[!tp]
\begin{center}
\includegraphics[width=0.5\textwidth]{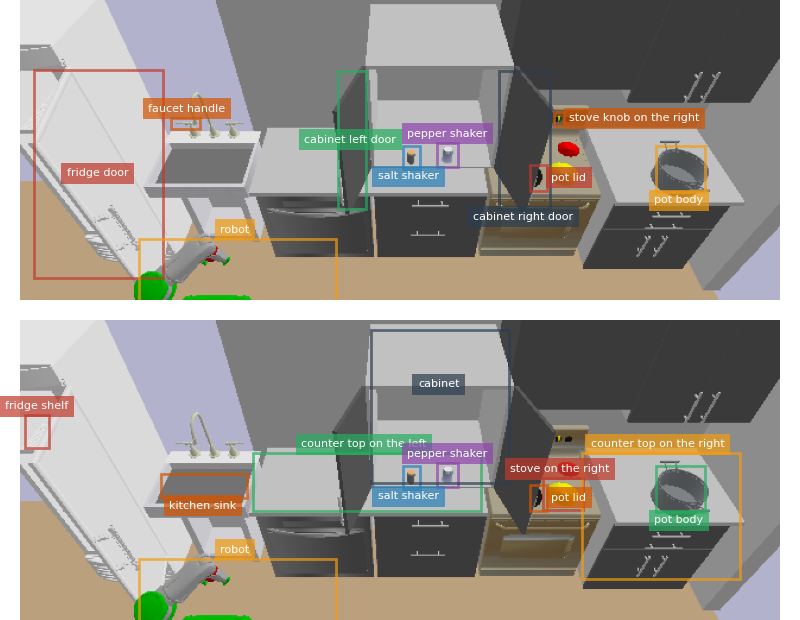}
\end{center}
\caption{\small An example input image to the VLM, which are annotated with object names and bounding boxes. The top image marks movable objects and articulated joints, while the bottom image marks movable objects and placement surfaces. }\vspace{-8pt}
\label{fig:query-image}
\end{figure}

We assume that problems are represented as correct 3D geometric model of the world plus the robot and a natural language goal, e.g. ``make chicken soup''. Because the geometric model of the world cannot be fully communicated to a VLM with image renders of the scene alone, we provide the VLM a text description of the scene that lists the objects and the relations they satisfy. For example, Figure~\ref{fig:query-image} is generated by rendering the scene in the PyBullet~\cite{coumans2019} simulator and labeling the observable objects using ground-truth semantic segmentation. This can also be generated with semantic segmentation models for real-world experiments.

The system outputs a sequence of robot commands (joint angle trajectories) to satisfy each subproblem identified by the VLM and verified by the TAMP planner.  The resulting state after achieving a subgoal is the initial state for planning and execution of the next subgoal.

\begin{figure}[!tp] \centering
\vspace{-7pt}
\begin{adjustwidth}{6pt}{1.7em}
\begin{lstlisting}[style=askvlm]
Plan a short sequence of [OUTPUT] that accomplishes the following {goal}. 
[RESPOND_WITH] where <obj>, <surface>, <joint>, <button>, <handle> must be items from the following {planning_objects}.
Currently, you can see {object_relations}. The accompanying image ... {img}
{action_history} {failed_action} {collided_objects}
\end{lstlisting}
\end{adjustwidth}
\vspace{-7pt}\hspace{0pt}
\begin{minipage}{0.12\textwidth}
\scriptsize a) The initial question to the VLM. Text formatted using \texttt{\textbf{[\textcolor{YangRed}{RED}]}} differs when asking for {\em subgoals} versus {\em actions}.
\end{minipage}\hfill
\begin{minipage}{0.35\textwidth}
\begin{adjustwidth}{0.5em}{0.5em}
\begin{lstlisting}[style=vlmanswer]
Open the fridge door to access the chicken leg.
Pick up the chicken leg from the fridge shelf ...
\end{lstlisting}
\end{adjustwidth}
\end{minipage}

\vspace{-2pt}
\begin{adjustwidth}{6pt}{1.7em}
\begin{lstlisting}[style=askvlm]
Translate the above intermediate goals into a formal language defined by the following subgoals.
subgoals = ['On(<movable>, <surface>)': the result of picking up <movable> then placing it on <surface>, ...] %
Please answer with objects in the respective types: {objects_by_types}
\end{lstlisting}
\end{adjustwidth}
\vspace{-7pt}\hspace{0pt}
\begin{minipage}{0.20\textwidth}
\scriptsize b) The VLM response and our next question when asking for {\em subgoals}.
\end{minipage}\hfill
\begin{minipage}{0.27\textwidth}
\begin{adjustwidth}{0.5em}{0.5em}
\begin{lstlisting}[style=vlmanswer]
Opened(fridge door), 
Picked(chicken leg), ...
\end{lstlisting}
\end{adjustwidth}
\end{minipage}

\vspace{-2pt}
\begin{adjustwidth}{6pt}{1.7em}
\begin{lstlisting}[style=askvlm]
Translate the each of the listed actions in English into a formal language defined by the following primitive actions. Each action in English may correspond to multiple actions:
actions =['pick(<obj>)': it contains one argument. The robot must have an empty hand to pick up an object. ...] %
Please answer with objects in the respective types: {objects_by_types}
\end{lstlisting}
\end{adjustwidth}
\vspace{-7pt}\hspace{0pt}
\begin{minipage}{0.20\textwidth}
\scriptsize c) The VLM response and our next question when asking for {\em actions}.
\end{minipage}\hfill
\begin{minipage}{0.27\textwidth}
\begin{adjustwidth}{0.5em}{0.5em}
\begin{lstlisting}[style=vlmanswer]
open(fridge door), 
pick(chicken leg), ...
\end{lstlisting}
\end{adjustwidth}
\end{minipage}

\caption{\footnotesize Conversation Template for querying VLMs and example responses used by \ours (ab) and baseline VLM + Motion Planning (ac). The same templates are used during reprompting, with text formatted using \texttt{\textbf{\{\textcolor{YangPurple}{Purple}\}}} representing updated information.
}\vspace{-8pt}
\label{fig:query}
\end{figure}

\subsection{Approach}\label{sec:approach}

Figure~\ref{fig:algorithm} shows an overview of our approach, and Algorithm~\ref{alg:algorithm} shows the corresponding pseudocode.  
There are \textit{two} distinct modes of operation of the system depending on what the VLM is prompted to produce:
\begin{itemize}
    \item Predicting {\em subgoals}: the VLM produces subgoals in PDDL format, using on a list of provided predicates %
    \item Predicting {\em actions}: the VLM produces a sequence of high-level actions, drawing from a set of legal actions.
\end{itemize}
After checking the semantic consistency of the subgoals or the actions, a problem is constructed for the TAMP planner.  
In the subgoal setting, the TAMP planner must construct a detailed motion plan that may involve moving multiple objects, opening cabinets, etc.  In the action setting, the TAMP planner still needs to convert the high-level action to actual robot commands with continuous parameters for grasps, paths, etc.  There may be failures at several points in this process which are addressed by reprompting the VLM.

{\footnotesize
\begin{algorithm}[!tp]  %
  \caption{ \ours}  \label{alg:algorithm}
  \small
  \begin{algorithmic}[1]
    \Require $\gO, \gI, \gG^{\text{eng}}, \textit{flag}_{a(actions)}, N_{\text{reprompt}}, N_{\text{TAMP}}$
    \State $\textcolor{YangGreen}{\pi \leftarrow [\,]}$ %
    \State $\textcolor{YangGreen}{\gI, \gi \leftarrow \proc{observe-state-and-image}()}$
    \State $\textcolor{YangGreen}{\proc{vlm} \leftarrow \proc{query-fn-gen}(\gO, \gG^{\text{eng}}, \textit{flag}_{a}, N_{\text{reprompt}})}$
    \State $\textcolor{YangGreen}{\bm{\gG} \leftarrow \proc{vlm-query}(\gI, \gi, \pi)}$%
    
    \While{$\textcolor{YangBlue}{\proc{len}(\bm{\gG}) > 0}$}
      \State $\textcolor{YangBlue}{\gG_k \leftarrow \bm{\gG}.\textit{pop}()}$  %
      \State $\textcolor{YangBlue}{\gO', \gO_c \leftarrow \proc{reduce-object}(\gO, \gI, \gG_k )}$\ %
      \State $\textcolor{YangBlue}{r \leftarrow \proc{check-semantics}(\gG_k)}$ %
      \If{$\textcolor{YangBlue}{r = \proc{succcess}}$}
          \State $\textcolor{YangBlue}{r, \pi_k, \tau_k, \gO_c \leftarrow \proc{tamp}(\gO', \gI, \gG_k, N_{\text{TAMP}})}$ %
          \If{$\textcolor{YangBlue}{r = \proc{Success}}$}
            \State $\textcolor{YangBlue}{\pi.\textit{extend}(\pi_k)}$
            \State $\textcolor{YangBlue}{\gI \leftarrow \proc{execute}(\tau_k)}$
            \State $\textcolor{YangBlue}{\gI, \gi \leftarrow \proc{observe-state-and-image}()}$ %
          \EndIf
      \EndIf
      \If{$\textcolor{YangOrange}{r = \proc{Failure}}$}
        \State $\textcolor{YangOrange}{\bm{\gG} \leftarrow \proc{vlm}(\gI, \gi, \pi, \gO_c)}$%
      \EndIf
    \EndWhile
  \end{algorithmic}
\end{algorithm}
}

\subsection{Using VLM for Subgoal or Action Sequencing}\label{sec:vlm}

Each VLM query has two phases. First, the VLM takes in an English description of the goal $\gG^{eng}$, all the objects in the scene $\gO^{eng}$, and their spatial relations $\gI^{eng}$. 
It outputs a sequence of intermediate subgoals $\{\gG^{eng}_i\}_{i=1}^K$ or actions $\hat{\pi}^{eng}$ in English, as shown in Figure~\ref{fig:query}a. Next, it takes a description of the possible subgoals or actions to translate the English answer into.
It outputs a sequence of intermediate subgoals $\{\gG_i\}_{i=1}^K$ (Figure~\ref{fig:query}b) or actions $\hat{\pi}$ (Figure~\ref{fig:query}c) in PDDL format that the TAMP planner consumes. 

\paragraph{Predicting Subgoals} After predicting subgoals in English $\{\gG^{eng}_i\}_{i=1}^K$, the VLM is prompted to translate them into a sequence of goal tuples $\{\gG_i\}_{i=1}^K$ (a single grounded predicate), given a pre-defined list of predicates $\gP$ along with the types of objects allowed for each position\footnote{For example, \texttt{\footnotesize On$\langle$object, surface$\rangle$}, \texttt{\footnotesize Sprinkled$\langle$object, object$\rangle$}, \texttt{\footnotesize Opened$\langle$joint$\rangle$}, and \texttt{\footnotesize TurnedOn$\langle$joint$\rangle$}.}. 
We currently use a single goal tuple for each subgoal, but this could be extended to a conjunction of goal literals.

\paragraph{Predicting Actions}  After predicting actions in English, the VLM translates the plan skeleton $\hat{\pi}^{eng}$ into a sequence of actions $\{\hat{a}_i\}_{i=1}^K$, given a pre-defined list of actions $\gA$\footnote{For example, \texttt{\footnotesize pick$\langle$object$\rangle$}, \texttt{\footnotesize place$\langle$object, region$\rangle$}, \texttt{\footnotesize sprinkle$\langle$object, region$\rangle$}, and \texttt{\footnotesize open$\langle$joint$\rangle$}.}, along with English descriptions of the preconditions of applying the action. Figure~\ref{fig:query}c) lists an example precondition that a ``hand should be empty before picking an object''. 
These preconditions are used when prompting the VLM in order to improve reasoning accuracy.

The VLM may produce goal literals or actions with semantic errors, e.g. using objects with the wrong type or objects that don't exist in the world, or using the wrong number of arguments. Each entry in the VLM output is verified with pure symbolic planning in a simplified PDDL formulation that only specifies the discrete arguments for predicates and actions.  
When one entry is found to be impossible, an error message is fed back to VLM to generate subgoals or actions again. The algorithm returns failure when a maximum number of queries $N_{\text{reprompt}}$ is reached.

\subsection{Using TAMP to Refine Subgoals or Action Sequences}\label{sec:tamp}

Given a sequence of subgoal or actions, the TAMP system refines each one in order, generating a sequence of grounded action plans and corresponding motion trajectories. 

\paragraph{TAMP problems} We represent TAMP problems using an extension of the Planning Domain Definition Language (PDDL)~\cite{mcdermott1998pddl}, a logic-based action language, that supports planning with continuous values~\cite{garrett2020PDDLStream}. We define a TAMP {\em domain} $\gD = \langle \gP, \gA\rangle$ by a set of {\em predicates} $\gP$ and {\em actions} $\gA$. {\em Predicates} and {\em actions} can be represented as tuples consisting of a name and a list of typed arguments. The arguments may be (1) discrete, such as object and part names, or (2) continuous, such as object poses, object grasps, robot configurations, object joint angles, and robot trajectories. %

We define a TAMP {\em problem} $\langle \gO, \gI, \gG, \gD\rangle$ using a set of {\em objects} $\gO$ (constants specific to the problem), a set of {\em initial literals} $\gI$, a conjunctive set of {\em goal literals} $\gG$, and the planning domain $\gD$. A {\em literal} is a predicate with an assignment of values to its arguments. The set of initial literals defines a state of the world. Each {\em grounded action} defines a deterministic transition of the world state. 

A {\em solution} $\pi$ is a finite sequence of grounded action instances that, when sequentially applied to the initial state $\gI$, produces a terminal state where the goal literals $\gG$ all hold. %
A {\it plan skeleton} $\hat{\pi}$ is a sequence of partially grounded actions, where the discrete parameters are bound but the continuous parameters are unbound.

\begin{figure}[!tp]
\begin{center}\vspace{-3px}
\includegraphics[width=0.6\textwidth,right]{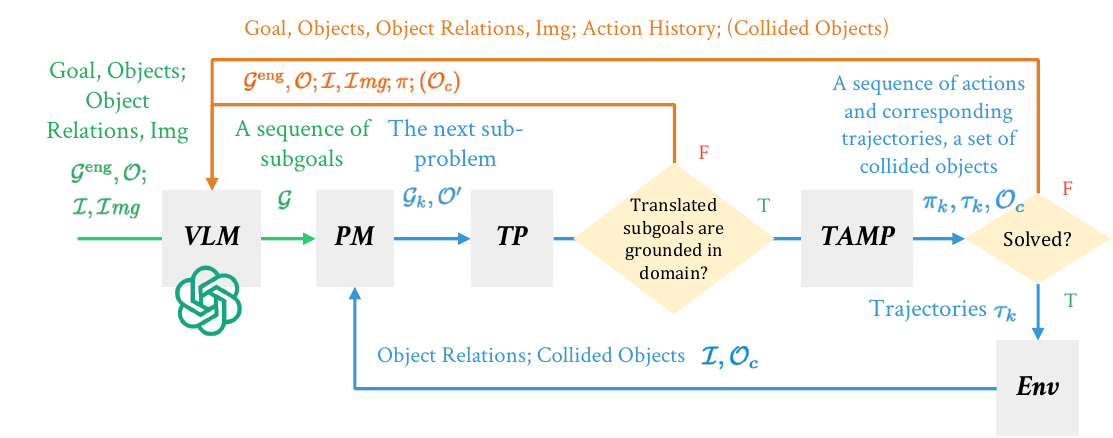}
\end{center} %
\caption{Our \ours Algorithm. \textit{\textbf{PM}} means Problem Manager, which formulates the next TAMP sub-problem to solve. \textit{\textbf{TP}} means Task Planning, which checks the semantics of subgoals.}
\label{fig:algorithm}\vspace{-8pt}
\end{figure}

\paragraph{Planning for Subgoals} For each goal literal $\gG_k$, a {\em subproblem} $\langle \gO_k, \gI_k, \gG_k\rangle$ is generated, which TAMP takes in then returns solution $\langle \pi_k, \tau_k, \gO_c\rangle$ if successful and $\langle r=\proc{Failure}, \gO_c\rangle$.  $\gI_k$ is the current state and involves all objects. $\gO_k$ is a small subset of objects to be considered for grounding predicates and actions during planning.  Reducing the universe of objects reduces the size of the action space (and thus branching factor) of this subproblem. 
First, planning is attempted with only the objects that are mentioned in the goal or apart of the robot state (i.e. currently grasped). During planning, the system records $\gO_c$, the movable or articulated objects that collide with the robot in some future state (e.g. while sampling object grasps, solving inverse kinematics, or planning motion). If planning with the goal-relevant objects fails, the often sparse subset of objects that contributed to collisions is added to the set of objects and the TAMP planner is called again. 
The subgoal is declared unreachable after $N_{\text{TAMP}}$ unsuccessful calls to the TAMP planner. When planning is successful, the system execute the motion trajectories $\tau_k$ and extend grounded plan $\pi_k$ to the whole history of actions executed. It then observe the environment and obtain updated $\gI_{k+1}$.

In this formulation, omitting objects from $\gO$ generally reduces the set of changes the robot can make in the domain but does not, for example, remove objects from the world that might cause collisions.
Limiting the relevant objects is most effective when the goal predicates are designed to be the effects of actions in the domain. For example, \texttt{\small Heat$\langle$object$\rangle$} is uninformative as it leaves out arguments like the heating surface and appliance handle; planning without those objects will fail. To avoid this, all predicates we ask the VLM to generate are on the spatial and motion level, which specify the directly relevant entities, such as \texttt{\small PlaceOn$\langle$object, surface$\rangle$} and \texttt{\small TurnHandle$\langle$joint$\rangle$}.

\paragraph{Refining Actions} Given a partially grounded action $a_k$, which includes only discrete parameters such as objects and robot arms, the system first finds its symbolic effects and uses them as $\gG_k$. 
Given a subproblem $\langle \gO_k, \gI_k, \gG_k\rangle$, the TAMP planner is called in a manner where it is constrained to use the partial plan skeleton $\hat{\pi}_k = [a_k]$.
This forces the planner to find a plan that uses that action; however, the planner still needs to refine the action by choosing values for its continuous parameters, and the planner may need to add additional actions, for example a base motion to reach the target action.
This variant also uses reduced objects and is allowed multiple trials, as in general subgoal planning.

\subsection{VLM Replanning after TAMP Failure} \label{sec:replanning}

When the TAMP planner fails to solve a subgoal or action after $N_{\text{TAMP}}$ trials, the VLM is prompted again as described in Algorithm~\ref{fig:algorithm} and Figure~\ref{fig:algorithm}. It takes in an updated description of the scene, the sequence of actions already executed, and collision objects detected during failed runs. The VLM reprompting is carried out at most $N_{\text{reprompt}}$ times before the system declares failure. %
Our approach assumes that there are no long-term low-level dependencies among the subgoals that can {\em only} be addressed by making one long detailed plan for the entire problem. In other words, as long as it is not possible to become irreversibly stuck after taking some actions, we can factor the whole problem into smaller ones, counting on plans for the later subgoals to resolve geometric difficulties caused by the previous ones.

\section{Experiments}
\label{sec:experiments}

\begin{figure*}[!tp]
\begin{center}
\includegraphics[width=1\textwidth]{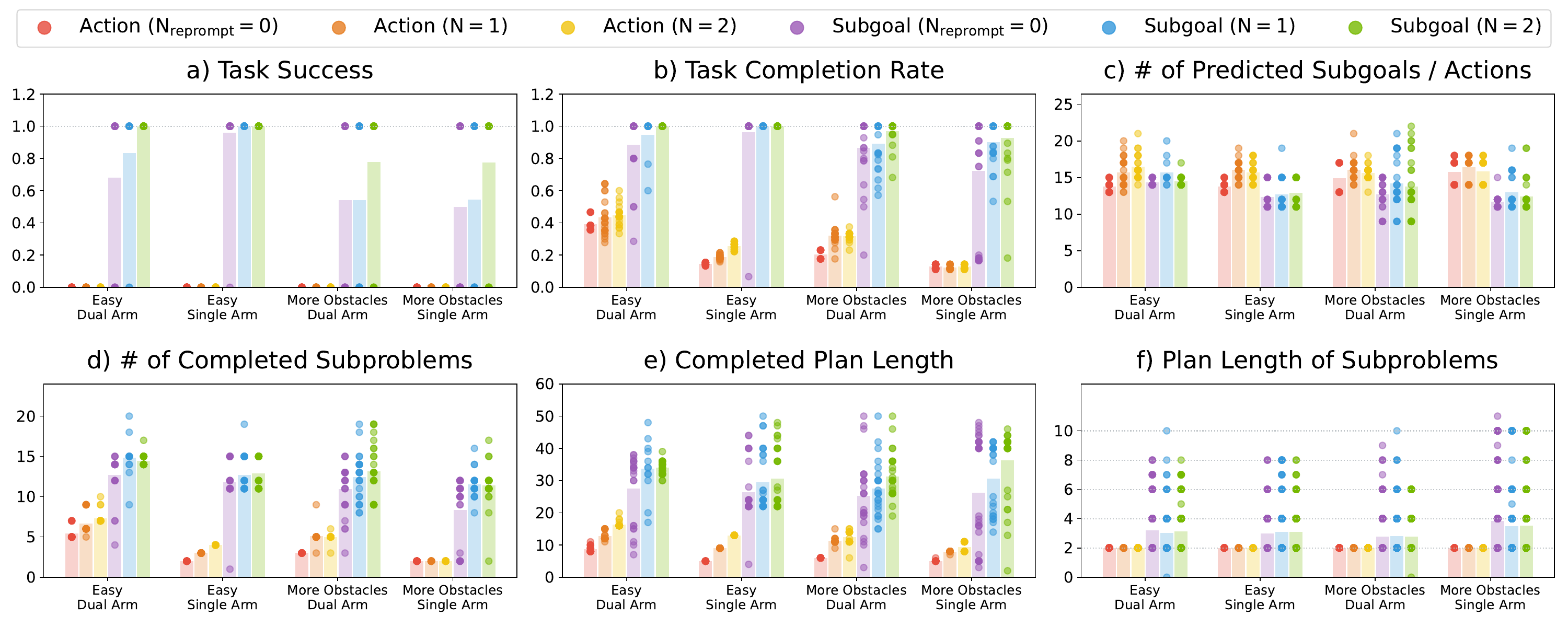}
\end{center} %
\caption{
Our experimental results show that that 1) predicting subgoals (\ours) outperforms predicting actions, 2) reprompting helps when subgoals (\ours) as number of reprompt tries increases but not when predicting actions. All six methods are run for 30 random trials on four problem difficulties, with increasing numbers controllable robot arms and manipulable obstacles.}
\label{fig:results}
\end{figure*}

\noindent We run experiments to answer the following questions:
\begin{enumerate}
    \item Which mode of VLM sequencing gives better task completion performance: predicting subgoals or actions?
    \item What's the extent to which reprompting improves performance? Does increasing the compute budgets increase the number of problems \ours{} can solve?
\end{enumerate}

\subsection{Baselines and Ablations}

We compare two approaches, each with three variants that allow VLM reprompting for $N_{\text{reprompt}} \in \{0, 1, 2\}$ times:
\begin{itemize}
    \item \textit{VLM Subgoal Sequencing + TAMP refinement (ours)} uses VLMs to generate a sequence of subgoals which a TAMP planner solves. 
    \item \textit{VLM Action Sequencing + limited TAMP refinement} uses VLMs to generate actions which a TAMP planner refines, but where no additional actions can be added other than moving the robot's base. This baseline is representative of \cite{zhang2023grounding,skreta2024replan,wang2024llm,joublin2023copal}.
\end{itemize}

\subsection{Task Suite and Robot Embodiment}

We consider the task of making chicken soup in a kitchen with 5 movable objects (e.g. food and seasoning), 8 surfaces (e.g. counter, stove burners, sink, pot), 2 spaces (enclosed in doors or a drawer), and 6 articulated objects (e.g. doors, knobs). In the \textit{Easy} case, all doors are initially open and the pot lid is on the counter. In the \textit{More Obstacles} case, all doors are initially closed and the pot lid is covering the pot body. To ensure that generated problems are feasible, we set objects to fixed initial poses and doors to slightly different open positions. Note that the subsequent object poses and joint positions are randomly sampled during planning, so the induced subproblems vary drastically after the initial state. 

The robot can change the pose of objects via pick and place actions as well as change the joint positions of articulated joints through pulling, pushing, and rotating its wrist.
In the \textit{Single-Arm} case, the robot is allowed to use only its left arm, while the \textit{Dual-Arm} case allows it to use both arms. Note that this kitchen environment is quite challenging, e.g., the sink is small and the fridge door and faucet allow a very limited range for the base and arm to position the  pot in the sink without collision. 

Altogether, we compare six methods on four variations of the task. Each method is run for 30 random trials.  For each task, we ask the VLM for five plans and we sample randomly from those for each of the 30 repetitions.
We measure the following performance metrics:
\begin{itemize}
    \item \textit{Task Success}. The algorithm successfully refines all subgoals or actions generated by the VLM into collision-free motion trajectories. 
    \item \textit{Task Completion Percentage}. The number of sub-problems solved out of all subproblems generated by the VLM and the Problem Manager, including completed problems and unfinished problems from the last query.
\end{itemize}

\subsection{Implementation Details}

We use \texttt{gpt-4o-mini} as our VLM ~\cite{achiam2023gpt}, with temperature $= 0.2$. For TAMP, we use the diverse planning mode of PDDLStream (as in \cite{yang2023piginet} but without the task plan feasibility predictor) with a maximum number of considered plan skeletons equal to $12$. For action refinement, we use the same planner but constrain it to include only the predicted action and moving the base in the plan skeleton. We use $N_{\text{TAMP}} = 3$ as the number of planning runs allowed for each subproblem, with increasing number of world objects included in planning. In other words, if the planner fails in the first two runs due to a sampling failure or not including a sufficient set of world objects but succeeds the last time, we still count the subproblem as a success.

\subsection{Results}

\paragraph{Predicting Subgoals Significantly Outperforms Predicting Actions}

Compared to to using the VLM for predicting actions that has zero success rate across all variations of the problems, \ours succeeds 50 to 100\% of the time as shown in Figure~\ref{fig:results}a). The full TAMP planner successfully fills in the actions to resolve geometric infeasibility by moving articulated or movable obstacles (using on average 1 to 2 actions as shown in Figure~\ref{fig:results}f). When comparing task completion percentage, it also significantly outperforms baselines (72 to 100\% versus 15 to 45\%). More subproblems are solved (Figure~\ref{fig:results}d) out of similar length of subproblem sequences proposed (Figure~\ref{fig:results}c) and the resulting plans have more actions (Figure~\ref{fig:results}e). See the supplementary video for example execution traces generated by \ours.
 
\ours visibly benefit from VLM reprompting, with task success increasing by 47 to 55\% on the harder problems, as shown in Figure~\ref{fig:results}a. As the number of reprompting runs increases, the performance of the subgoal variants increases as it gives TAMP more tries to solve the geometrically difficult subproblems and it may inject intermediate subgoals that make it easier to achieve the failed subgoal. As seen from individual run statistics, reprompting reduced the variance in the completion percentage. In comparison, reprompting didn't help the variant that generates actions, even though the prompt includes action history, collision summary, and description of which robot arms are holding which objects. This shows that the VLM cannot be relied upon to consider long-horizon history, complex world description, and geometric infeasibility when predicting actions.

\section{Failure Analysis}

We overview several system failure modes.
In particular, VLMs are fairly experimental and  routinely make inaccurate predictions, although we expect them to improve over time.

\subsection{VLM Failures}

\paragraph{Subgoal/Action translation to PDDL} The second phase of VLM query process asks the VLM to translate English answers to formal PDDL language of goal or action tuples. The VLM may 1) miss certain subgoals or actions when one sense contains multiple subgoals or actions, and 2) generate a tuple with the wrong number of arguments or misspell the object name. %

\paragraph{Infeasible actions} The subproblem may not be solvable because there exist movable or articulated objects (e.g., robot is asked to turn on the stove but the pot is placed in a location that blocks any path to grasp the knob handle), or because the precondition is not met (e.g. robot asked to close a door but the robot's only arm is holding an object)\footnote{Although the VLM is given a rule in the prompt that says ``\texttt{\scriptsize You must have at least one empty hand before you can pick up an object or open or close a joint}'', it still sometimes generates an English goal ``\texttt{\scriptsize Close the cabinet left door with the hand that was holding the salt shaker}'' and then translates it into ``\texttt{\scriptsize closed-door(cabinet left door)}'' which is infeasible.}.  %

\subsection{TAMP Failures}

\paragraph{Didn't find a feasible plan skeleton} The planner is allowed to try 12 plan skeletons in order of increasing task length. For a subproblem that involves six planning objects and two obstacles to clear out, the correct plan skeleton may not be ranked in the first 12 allowed. %

\paragraph{Found feasible plan skeletons but failed to refine them within the compute budget} Each low-level samplers (e.g., pose sampler, grasp sampler, inverse-kinematic solver) is allowed a limited number of attempts. During refinement, they are attempted in sequence and need all be successful in order to refine the plan skeleton. %

\section{Discussion} 
\label{sec:discussion}

We present \ours, a system for solving long-horizon manipulation planning problems by marrying the strengths of VLM's language understanding and commonsense reasoning abilities, to the strengths of TAMP's ability to find feasible task skeletons and generate collision-free trajectories that respect all geometric constraints. The combined system effectively overcomes the shortcomings of 1) VLM's lack of geometric and long-horizon reasoning abilities by letting TAMP fill in required actions and parameter values and 2) TAMP's explosive computational complexity by leveraging a VLM to break down the problem along with allowing it to correct for failures on a short horizon via reprompting.

Future work involves training dexterous dual-arm manipulation policies to achieve subgoals given visual inputs.

\section*{Acknowledgment}
We gratefully acknowledge support from AI Singapore AISG2-RP-2020-016; from NSF grant 2214177; from AFOSR grant FA9550-22-1-0249; from ONR MURI grant N00014-22-1-2740; from ARO grant W911NF-23-1-0034; from the MIT Quest for Intelligence; and from the Artificial Intelligence Institute. This work was partially conducted during an internship at NVIDIA Research.

\bibliographystyle{IEEEtran}
\bibliography{references}

\end{document}